\documentclass[10pt,twocolumn,letterpaper]{article}

\usepackage[pagenumbers]{wacv} %

\def\cleanversion{}
\ifdefined\cleanversion
  
  \newcommand{\del}[1]{}
  
  \newcommand{\delfloat}[1]{}
\else
  
  \newcommand{\del}[1]{{\color{red}#1}}
  \makeatletter
  \gdef\nextfloatcolor{}
  \g@addto@macro\@floatboxreset{\nextfloatcolor\gdef\nextfloatcolor{}}
  \makeatother
  
  \newcommand{\delfloat}[1]{\gdef\nextfloatcolor{\color{red}}\input{#1}}
\fi

\usepackage{multirow}

\definecolor{wacvblue}{rgb}{0.21,0.49,0.74}
\usepackage[pagebackref,breaklinks,colorlinks,allcolors=wacvblue]{hyperref}

\def\wacvPaperID{2175} %
\def\confName{WACV}
\def\confYear{2027}

\title{LC-Flow: Learning Local Continuous Optical Flow and Confidence from Events}

\author{
Gunwoo Jeon$^{1}$ \quad Chaesong Park$^{1}$ \quad Jongwoo Lim$^{1,2}$\\
IPAI$^{1}$ \& ME$^{2}$, Seoul National University\\
{\tt\small \{gunw0, chase121, jongwoo.lim\}@snu.ac.kr}
}

\begin{document}
\maketitle

\begin{abstract}
Event cameras capture brightness changes with microsecond resolution, yet existing optical flow methods fail to fully exploit this temporal detail. Frame-based approaches accumulate events over a fixed window long enough to be densely populated, then regress a global flow field; they are accurate but tie their output to this training-time window and overfit to global scene priors that transfer poorly across domains, while model-based local methods operate statelessly, discarding temporal history between predictions and yielding inaccurate flows.
We propose \textbf{LC-Flow}, a local recurrent framework that incrementally processes event streams while maintaining independent hidden states for spatially localized regions. Unlike fixed-window processing, the persistent recurrent states accumulate temporal information across inference intervals, enabling flow estimation at arbitrary query timestamps. To address local motion ambiguity, LC-Flow jointly estimates flow and confidence, which enables filtering unreliable predictions and guides multi-scale aggregation into a dense flow field. By avoiding globally aggregated scene features, LC-Flow reduces reliance on scene-specific spatial priors. Trained on a single outdoor sequence, LC-Flow achieves state-of-the-art performance on MVSEC, including unseen indoor scenes, while additional zero-shot experiments on DSEC demonstrate the transferability of flow structure under scale-aligned evaluation.
\end{abstract}

\section{Introduction}
\label{sec:intro}

Event cameras are bio-inspired sensors that asynchronously capture per-pixel brightness changes with microsecond resolution and high dynamic range, making them well suited to high-speed motion estimation in challenging environments. Applications such as visual odometry (VO) and target tracking further demand motion estimates produced with low latency and at a high rate, for which local processing is well suited. However, accurate optical flow estimation from local observations alone remains challenging, and existing methods fall into two polarized paradigms: dense learning-based approaches and sparse local approaches.

Dense methods~\cite{eraft,liu2025edcflow} accumulate sparse events into synchronous representations (e.g., voxel grids) to leverage powerful deep neural networks. While accurate within their training domain, these representations must accumulate enough events to be informative, which imposes an accumulation window with its attendant latency and ties the prediction to the fixed rate used in training; their reliance on global spatial priors further makes them prone to domain overfitting.
Conversely, local methods~\cite{akolkar2020real,nagata2023tangentially,yuan2024learning} estimate motion directly from localized patches. However, many existing local approaches lack persistent temporal memory and rely on simplified geometric assumptions or handcrafted priors, making them susceptible to the aperture problem and event noise.
To bridge this gap, we propose LC-Flow, a local learning-based recurrent framework that estimates flow and confidence directly from local event streams. Unlike frame-based networks, we avoid global context aggregation to prevent domain overfitting. Instead, we introduce a Continuous Local Recurrent Network that assigns an independent, persistent hidden state to each local grid, whose state is updated incrementally as every event arrives. Following the continuous-time view of Gehrig \etal~\cite{gehrig2024dense}, the state can be queried at an arbitrary timestamp.

The sparse local flow field is the focus of this work: it is directly usable for downstream tasks such as VO and tracking, where reliable motion cues at active pixels suffice. However, purely local patches suffer from the aperture problem, where motion along uniform edges is ambiguous. We therefore jointly learn a confidence score that quantifies the reliability of each grid's prediction from its local spatiotemporal evidence.
Our localized estimates are not limited to sparse outputs: a multi-scale confidence-guided aggregation propagates reliable motion cues to uncertain regions, reconstructing a globally consistent flow field without global feature aggregation. On MVSEC, LC-Flow achieves state-of-the-art performance among evaluated local methods, while its dense aggregation achieves the lowest EPE across the evaluated sequences. Notably, despite being trained on a single outdoor sequence, LC-Flow maintains strong performance on unseen indoor scenes, demonstrating generalization across different scene conditions within MVSEC. Additional zero-shot evaluation on DSEC provides evidence of cross-dataset transfer in flow direction and spatial structure under scale-aligned evaluation.
Our main contributions are summarized as follows:
\begin{itemize}[leftmargin=*, nosep]
\item \textbf{Local recurrent architecture with continuous-time queries:} We propose a learning-based local recurrent framework that maintains independent hidden states for spatially localized event streams, enabling incremental temporal modeling and flow queries at arbitrary timestamps.
\item \textbf{Confidence-aware local flow estimation and aggregation:} We jointly learn optical flow and confidence from local event observations. The predicted confidence identifies reliable motion estimates and guides multi-scale aggregation to reconstruct dense flow fields.
\item \textbf{State-of-the-art result on MVSEC:} LC-Flow achieves state-of-the-art performance on MVSEC, including indoor sequences unseen during outdoor-only training. Additional evaluation on DSEC provides evidence of zero-shot transfer of flow direction and spatial structure.
\end{itemize}

\section{Related Works}
\subsection{Dense Flow Estimation}
Dense optical flow estimation from event cameras has been extensively studied through both model-based and learning-based approaches. Model-based approaches estimate motion by explicitly modeling event generation and optimizing geometric consistency. Multi-CM~\cite{shiba2022secrets} reconstructs sharp edge structures through contrast maximization, while other methods exploit handcrafted spatio-temporal surfaces for matching-based motion inference~\cite{nagata2021optical,almatrafi2020distance}. However these can be brittle in complex scenes.

Learning-based dense methods bridge asynchrony by accumulating events into synchronous, frame-like representations for high-capacity Neural Networks: EV-FlowNet~\cite{zhu2018ev} uses event-count representations with photometric losses, while E-RAFT~\cite{eraft}, TMA~\cite{liu2023tma}, and EDCFlow~\cite{liu2025edcflow} adapt image-flow architectures to voxel grids. Although accurate within their training domain, these dense methods incur accumulation latency, are sensitive to inference frequency, and rely on globally aggregated features that generalize poorly across domains. A parallel line explores SNN-based models for low-latency dense inference~\cite{lee2020spike,cuadrado2023optical,zhang2023event}, although stabilizing training and achieving competitive accuracy remain challenging.

\subsection{Local Flow Estimation}
Beyond dense flow, several works estimate sparse motion directly from local event neighborhoods, targeting low latency and fine temporal resolution. Classical techniques such as Lucas--Kanade (LK)-based methods~\cite{barranco2014contour} compute motion from localized spatio-temporal neighborhoods but suffer from the aperture problem.
A number of methods aim to improve robustness and mitigate the aperture problem within this sparse local paradigm. ARMSflow~\cite{akolkar2020real} employs robust kinematic plane fitting, VecKMFlow~\cite{Veckmflow} predicts normal flow from compact event patches, and Shiba \etal~\cite{shiba2023fast} propose event-by-event triplet matching to align events across space and time. To share information across spatial neighborhoods, TEGBP~\cite{nagata2023tangentially} propagates tangentially elongated Gaussians via belief propagation; however, such approaches typically rely on explicit smoothness assumptions and handcrafted noise models, which can become brittle in complex real-world scenes. Crucially, none of these local methods maintain a learned persistent memory updated with incoming events. To our knowledge, LC-Flow is the first local learning-based method to maintain persistent hidden states per local grid, enabling continuous inference with accumulated temporal context.

Beyond benchmarking, local sparse flow is also valuable for downstream tasks such as visual odometry and target tracking, where reliable motion cues at active locations facilitate data association~\cite{wang2023lightweight, senst2012robust, liu2022robust}. This further motivates estimating prediction confidence for reliable deployment.

\subsection{Continuous-time and Streaming Representations for Event Data}
A growing body of work argues that the windowed-to-dense paradigm introduces latency and discards fine-grained temporal structure, and that models trained at a fixed accumulation rate generalize poorly to other inference frequencies~\cite{baldwin2021tore, lin2025compressed, zubic2024state, lu2026flexevent}. This has driven interest in continuous-time processing: sparse event-by-event computation via recurrent updates or message passing~\cite{sekikawa2019eventnet, schaefer2022aegnn}, state-space formulations with persistent latent states~\cite{zubic2024state, schone2024scalable}, and continuous-time motion fields that reduce discrete-time aliasing~\cite{wang2024continuous}.

Together, these trends motivate our design: rather than committing to a single fixed accumulation rate, we maintain persistent local hidden states and update them incrementally. This decouples inference from a fixed accumulation window, preserves the temporal ordering of events within each interval, and lets computation adapt to local scene activity. Our formulation is complementary to dense continuous-time flow~\cite{gehrig2024dense}, which accumulates events into a global representation over a window and parameterizes a continuous-time flow field from globally shared context. LC-Flow instead keeps a per-grid state driven only by local events and queries each grid at an arbitrary timestamp without global aggregation, targeting low-latency sparse flow with strong cross-domain transfer rather than high-accuracy dense reconstruction. %

\begin{figure*}[t]
    \centering
    \includegraphics[width=0.9\textwidth]{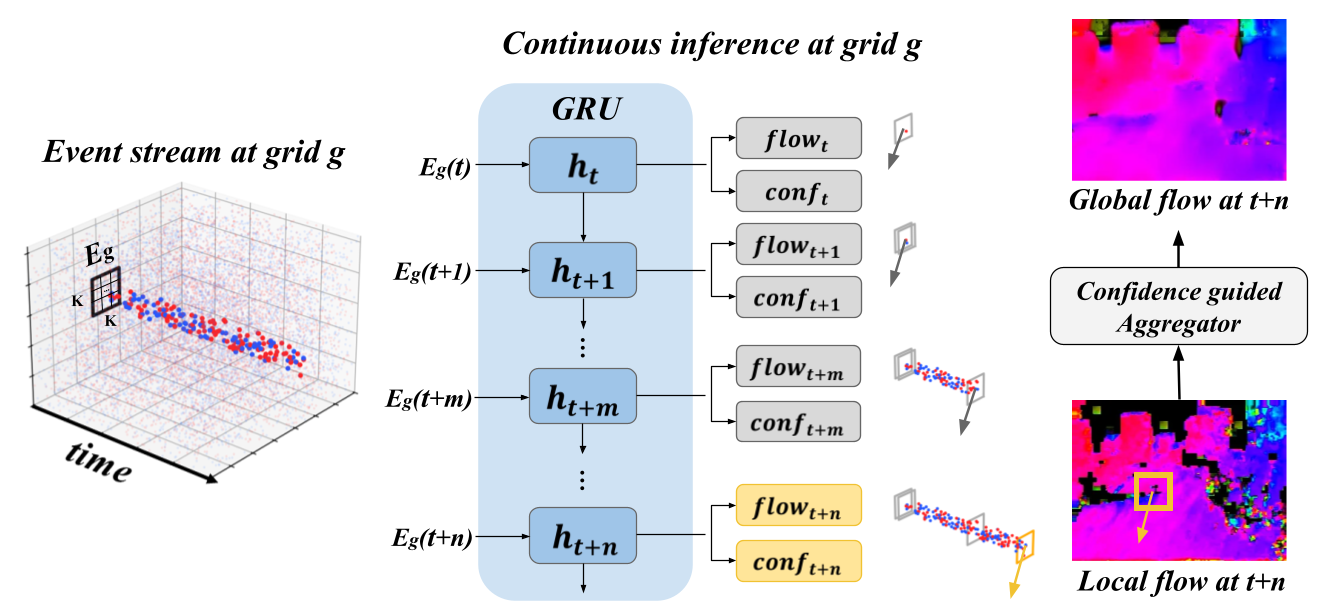}
    \caption{Overall architecture of LC-Flow. The Continuous Local Recurrent Network processes a localized event stream at grid $g$, where each event is embedded as $e_{\text{local}} = (2(x_i - x_c)/K,\; 2(y_i - y_c)/K,\; (t_i - t_{i-1}) \cdot \alpha,\; p_i)$, with $K$ and $\alpha$ denoting the spatial and temporal scaling factors, respectively. The GRU maintains a persistent hidden state $h_i$ per grid to continuously predict local flow and confidence. For global estimation, the sparse local predictions are integrated by the Confidence-guided Aggregator, which performs multi-scale confidence-weighted fusion across strided grid locations to reconstruct a globally consistent flow field.}

    \label{fig:architecture}
\end{figure*}

\section{Method}
\label{sec:method}
As illustrated in \cref{fig:architecture}, LC-Flow consists of (i) its core \emph{Continuous Local Recurrent Network}, which estimates flow and confidence purely from local events while maintaining a persistent hidden state for each local grid $g$, enabling event-driven flow querying at arbitrary timestamps from the per-grid event stream $\mathcal{E}_g$, and (ii) an optional \emph{Confidence-guided Aggregator} that reconstructs a global flow field from these sparse local predictions when a dense output is required.
\subsection{Continuous Local Recurrent Network}
\subsubsection{Problem Formulation}
Given a continuous stream of asynchronous events $\mathcal{E}=\{e_k\}_{k=1}^{N}$, our goal is to estimate, at any timestamp $t$, the optical flow $\mathbf{f}(x,y,t)=(u,v)$ and a confidence score $c(x,y,t)$ at pixel $(x,y)$. We define $\mathbf{f}(x,y,t)$ as the displacement over an interval $\Delta t$, where $\Delta t$ is set to match the dataset's short ground-truth annotation interval. Instead of waiting for a fixed accumulation window, LC-Flow updates its estimate incrementally as events arrive, integrating motion history into the hidden state.

\subsubsection{Event Embedding}
Each event is represented by its spatial location, timestamp, and polarity, $e_i=(x_i,y_i,t_i,p_i)$. We process events locally within a $K\times K$ neighborhood centered at grid location $(x_c,y_c)$, where $K$ is an odd integer. Spatial coordinates are normalized relative to the grid center, and temporal information is encoded using the inter-event interval. The local event embedding is defined as
\begin{equation}
e_{\text{local},i}=
\left(
\frac{2(x_i-x_c)}{K},\;
\frac{2(y_i-y_c)}{K},\;
(t_i-t_{i-1})\cdot \alpha,\;
p_i
\right),
\label{eq:event_embedding}
\end{equation}
where $\alpha$ scales the temporal component (we set $\alpha=100$ in practice so that typical inter-event intervals have a magnitude comparable to the normalized spatial terms). We initialize $t_0$ to the timestamp of the first event in each grid. This embedding maps local spatial coordinates to $[-1,1]$ and captures relative temporal dynamics rather than absolute time.

\subsubsection{Continuous GRU Update}
We utilize a Gated Recurrent Unit (GRU)~\cite{chung2014gru} to aggregate temporal information by maintaining a persistent hidden state per grid:
\begin{equation}
h_i=\mathrm{GRU}(e_{\text{local},i},\,h_{i-1}).
\end{equation}
The hidden state is zero-initialized and updated for every incoming event, accumulating motion context without fixed temporal boundaries. We adopt the GRU for its simplicity and training stability.

\subsubsection{Prediction Heads}
The hidden state $h_i$ is fed into two separate MLP heads. The flow head outputs a 2D flow vector at the grid center $(x_c,y_c)$, denoted $\mathbf{f}_{\text{local}}\in\mathbb{R}^2$. The confidence head outputs a scalar $c\in(0,1)$ via a sigmoid activation, representing the estimated reliability of $\mathbf{f}_{\text{local}}$ given the local spatiotemporal evidence.

\subsubsection{Loss}
We jointly optimize the flow and confidence heads using a confidence-weighted regression objective inspired by uncertainty-based weighting:
\begin{equation}
\mathcal{L}
= c\,\left\lVert \mathbf{f}_{\mathrm{pred}}-\mathbf{f}_{\mathrm{gt}}\right\rVert_2
-\lambda \log c,
\label{eq:loss}
\end{equation}
where $\mathbf{f}_{\mathrm{pred}}$ and $\mathbf{f}_{\mathrm{gt}}$ denote the predicted and ground-truth flow at the supervised location, $c$ is the predicted confidence, and $\lambda$ is a balancing coefficient. The first term penalizes the flow error weighted by confidence, while the second term regularizes $c$ to prevent the trivial solution of driving confidence toward zero.
This objective is a reparameterization of the uncertainty-based weighting of Kendall \etal~\cite{kendall2017uncertainties}: substituting the confidence $c\in(0,1)$ for the precision $1/(2\sigma^2)$ yields the same regularization structure. We use $c$ directly because its bounded range is convenient for filtering and confidence-weighted aggregation; we thus refer to it as a confidence (equivalently, low-uncertainty) score.

\subsubsection{Inference}
While the formulation updates per event (Eq.~\ref{eq:event_embedding}), at inference we process events in short micro-batched chunks for GPU efficiency; the hidden state persists across chunks and event order is preserved, so chunk size is a scheduling parameter rather than a semantic accumulation window. The state is reset only when the interval between successive events, or from the last event to the query time, exceeds $\Delta t$; in the latter case the grid has scarce evidence for the $\Delta t$ displacement, is not inferred, and is deferred to the aggregator.

\subsection{Confidence-guided Aggregator}
The primary output of LC-Flow is sparse local flow. We also provide a patch-based aggregator that reconstructs a dense global field from these sparse outputs when required, propagating reliable local motion to nearby uncertain or inactive regions without global feature aggregation.

\subsubsection{Gridwise Recurrent Processing with Spatial Stride}
We deploy the local estimator only at grid centers sampled with stride $s$. Each deployed grid maintains an independent recurrent hidden state updated solely by events within its $K\times K$ spatial neighborhood, and there is no interaction across grids. Striding provides a direct trade-off between spatial coverage and computation.

\subsubsection{Multi-scale Confidence-guided Aggregation}
At each deployed grid center, LC-Flow predicts a local flow vector and confidence. Treating these strided predictions as samples on a regular grid, we bilinearly interpolate them to a full-resolution flow map $\tilde{\mathbf{f}}$ and confidence map $\tilde{c}$.
To aggregate context at multiple scales, for each scale $r\in\{1,2,4\}$ we
average-pool $\tilde{\mathbf{f}}$ and $\tilde{c}$ by a factor of $r$, yielding $\tilde{\mathbf{f}}_r$ and
$\tilde{c}_r$. We then compute a confidence-weighted local average over a
$k\times k$ neighborhood $\mathcal{N}_k(\mathbf{x})$ at each pixel $\mathbf{x}$:
\begin{equation}
\hat{\mathbf{f}}_r(\mathbf{x}) =
\frac{\sum_{q\in\mathcal{N}_k(\mathbf{x})} \tilde{c}_r(q)\,\tilde{\mathbf{f}}_r(q)}
{\sum_{q\in\mathcal{N}_k(\mathbf{x})} \tilde{c}_r(q) + \epsilon},
\end{equation}
where $\epsilon$ is a small constant for numerical stability, and bilinearly
upsample $\hat{\mathbf{f}}_r$ back to full resolution. A grid that is not inferred (gap $>\Delta t$) enters the aggregation with zero flow and zero confidence, so the confidence weighting gives it no weight and fills it from neighboring reliable grids.

To fuse information across scales without uniformly averaging, we adaptively predict per-scale fusion weights from the local multi-scale evidence. Specifically, we stack the flow (2 channels) and confidence (1 channel) from each scale within the $k\times k$ neighborhood into a local tensor $\mathcal{P}(\mathbf{x})\in\mathbb{R}^{k\times k\times 9}$. A lightweight patch-based CNN then predicts soft fusion weights:
\begin{equation}
[w_1(\mathbf{x}),\,w_2(\mathbf{x}),\,w_4(\mathbf{x})]
= \mathrm{Softmax}\!\left(\mathrm{CNN}(\mathcal{P}(\mathbf{x}))\right).
\end{equation}
The final aggregated flow is
\begin{equation}
\mathbf{f}_{\text{agg}}(\mathbf{x})
= w_1(\mathbf{x})\,\hat{\mathbf{f}}_1(\mathbf{x})
+ w_2(\mathbf{x})\,\hat{\mathbf{f}}_2(\mathbf{x})
+ w_4(\mathbf{x})\,\hat{\mathbf{f}}_4(\mathbf{x}).
\end{equation}

\subsection{Training Strategy}
\label{sec:trainingstrag}
\subsubsection{Training Data Construction}
We construct training samples from temporally sliced event streams provided by standard event-based optical flow benchmarks. Each slice is partitioned into local grids of size $K\times K$. A grid is considered valid if it contains at least 10 events, ensuring sufficient local motion evidence. For each valid grid, we extract its per-grid event sequence $\mathcal{E}_g$ and supervise the flow at the grid center with the slice-level ground-truth, mirroring our inference-time processing.

\subsubsection{Continuous Training Strategy}
Training on a single slice does not expose the recurrent state to the long temporal evolution required for continuous inference. We therefore form each training sample by concatenating $M$ consecutive slices ($M$ uniformly sampled from a fixed range), processing them sequentially to update the hidden state while supervising flow only at the final slice. This aligns training with event-driven inference and yields stable temporal representations over varying durations.

\subsubsection{Data Augmentation}
To improve generalization and robustness to local motion variations, we apply augmentation directly to the extracted local event streams. Geometric transforms are applied consistently to both event coordinates and ground-truth flow vectors:
\begin{itemize}
\item \textbf{Reflection.} Reflect events horizontally or vertically about the grid center, and transform flow accordingly.
\item \textbf{Rotation.} Rotate events around the grid center. We use both right-angle rotations ($90^\circ$, $180^\circ$, $270^\circ$) and random rotations sampled uniformly from $[0^\circ,360^\circ]$. For arbitrary angles, rotated coordinates are rounded to the nearest pixel to preserve the discrete grid structure.
\item \textbf{Polarity Reversal.} Invert event polarities to simulate contrast changes, leaving spatial and temporal coordinates unchanged.
\end{itemize}

\section{Experiments}
\label{sec:experiments}

\subsection{Experimental Setup}

\noindent\textbf{Datasets.}
We evaluate our method on MVSEC~\cite{zhu2018mvsec}, which provides ground truth at $\sim$22\,ms intervals (45\,Hz), matching the high-frequency, low-latency regime that LC-Flow targets. For MVSEC, the local flow model is trained exclusively on the \textit{outdoor\_day2} sequence and evaluated on \textit{outdoor\_day1} and \textit{indoor\_flying1,2,3}, following the standard evaluation protocol.  To probe cross-dataset transfer, we additionally evaluate the MVSEC-trained models on DSEC~\cite{gehrig2021dsec} without fine-tuning.

\noindent\textbf{Local flow model configuration.}
The image space is partitioned into local grids of size $15\times 15$ pixels, and each grid maintains an independent GRU with a hidden dimension of 256. The network is trained using the Adam optimizer with a batch size of 512 and a learning rate of $5\times 10^{-4}$. 

\noindent\textbf{Training schedule.}
The GRU, flow, and confidence heads are jointly trained for 450K iterations with the confidence balancing coefficient set to $\lambda=0.2$. Each training sample is constructed by concatenating $M$ consecutive temporal slices, where $M$ is uniformly sampled from $[1,10]$.

\noindent\textbf{Data augmentation.}
We use the augmentations described in~\cref{sec:trainingstrag}. A right-angle rotation ($90^\circ$ or $270^\circ$) is applied with 50\% probability. Additionally, polarity reversal, spatial reflection, or random rotation is sampled with a ratio of $2{:}3{:}3$, respectively.

\noindent\textbf{Confidence-guided Aggregator.}
We deploy the local model with stride $s=3$ and perform multi-scale aggregation with patch size $k=7$ and scales $\{1,2,4\}$. The local model weights are frozen during aggregator training. The aggregator is trained on the same \textit{outdoor\_day2} sequence for 50 epochs with a learning rate of $10^{-4}$, minimizing the endpoint error to ground truth over valid-GT pixels.

\begin{table*}[t]
\centering
\caption{Comparison with local flow methods on MVSEC. The first block uses the native MVSEC image frequency; the second uses $\Delta t = 0.05$s to match TEGBP's reported interval. \textbf{Bold} indicates the best performance; ``$-$'' denotes metrics not reported by the original work.}

\label{tab:unified_local_flow_final_v6}
\small
\begin{tabular}{ll ccc ccc ccc ccc}
\toprule
\multirow{2}{*}{Method} & \multirow{2}{*}{Type}
& \multicolumn{3}{c}{Outdoor1}
& \multicolumn{3}{c}{Indoor1}
& \multicolumn{3}{c}{Indoor2}
& \multicolumn{3}{c}{Indoor3} \\
\cmidrule(lr){3-5} \cmidrule(lr){6-8} \cmidrule(lr){9-11} \cmidrule(lr){12-14}
& & PEE($\downarrow$) & EPE($\downarrow$) & \%Out($\downarrow$)
& PEE & EPE & \%Out
& PEE & EPE & \%Out
& PEE & EPE & \%Out \\
\midrule

\multicolumn{14}{l}{Inference on MVSEC image frequency} \\
\midrule
Triplet~\cite{shiba2023fast} & MB & - & 0.90 & 3.1 & - & 1.10 & 2.9 & - & 1.70 & 13.4 & - & - & - \\
VecKM~\cite{Veckmflow} & SL & 0.88 & - & - & 0.97 & - & - & 1.09 & - & - & 1.04 & - & - \\
ARMS~\cite{akolkar2020real} & MB & - & 2.75 & - & - & 1.52 & - & - & 1.59 & - & - & 1.89 & - \\
\textbf{Ours} & SL & \textbf{0.19} & \textbf{0.24} & \textbf{0.05} & \textbf{0.32} & \textbf{0.48} & \textbf{0.20} & \textbf{0.40} & \textbf{0.62} & \textbf{0.58} & \textbf{0.37} & \textbf{0.58} & \textbf{0.49} \\

\midrule
\multicolumn{14}{l}{Inference on $\Delta t = 0.05$s} \\
\midrule
TEGBP~\cite{nagata2023tangentially} & MB & - & 1.46 & 11.1 & - & 1.14 & 6.25 & - & 1.87 & 16.4 & - & 1.54 & 11.8 \\
\textbf{Ours} & SL & - & \textbf{0.93} & \textbf{1.76} & - & \textbf{0.98} & \textbf{1.46} & - & \textbf{1.56} & \textbf{8.03} & - & \textbf{1.43} & \textbf{7.30} \\

\bottomrule
\end{tabular}
\end{table*}

\subsection{Evaluation Metrics}
We report three metrics. Endpoint Error (EPE) is the mean Euclidean distance between predicted and ground-truth flow over all valid pixels. Outlier Ratio (\%Out) is the percentage of valid pixels with EPE larger than 3 pixels. Projected Endpoint Error (PEE) follows the normal-flow evaluation of VecKMFlow~\cite{Veckmflow}, computing EPE after projecting the GT flow onto the direction of the predicted.

\begin{table*}[t]
\centering
\small

\caption{
Quantitative comparison on the MVSEC dataset. Avg.\ EPE and Avg.\ Out denote averages over all four sequences. \textbf{Bold} and \underline{underline} indicate the best and second-best results, respectively. Learning type: MB denotes model-based; USL/SSL/SL denote unsupervised/self-supervised/supervised learning.
}
\label{MVSECflow_table_withavg}
{%
\begin{tabular}{ll lclclclc cc}
\toprule
& \textbf{Method } 
& \multicolumn{2}{c}{\small Outdoor1}
& \multicolumn{2}{c}{\small Indoor1}
& \multicolumn{2}{c}{\small Indoor2}
& \multicolumn{2}{c}{\small Indoor3}
& \textbf{Avg.} & \textbf{Avg.} \\
& 
& \multicolumn{2}{c}{}
& \multicolumn{2}{c}{}
& \multicolumn{2}{c}{}
& \multicolumn{2}{c}{}
& \textbf{EPE} & \textbf{Out} \\
\cmidrule(lr){3-4} \cmidrule(lr){5-6} \cmidrule(lr){7-8} \cmidrule(lr){9-10}
\cmidrule(lr){11-11} \cmidrule(lr){12-12}
& ($dt=1$)
& EPE($\downarrow$) & \%Out($\downarrow$)
& EPE & \%Out
& EPE & \%Out
& EPE & \%Out
&  &  \\
\midrule

MB & \small Brebion et al. \cite{brebion2021real} 
& 0.53 & 0.20 & 0.52 & \underline{0.10} & 0.98 & 5.50 & 0.71 & 2.10
& 0.69 & 1.98 \\

& \small Nagata et al. \cite{nagata2021optical} 
& 0.77 & - & 0.62 & - & 0.93 & - & 0.84 & -
& 0.79 & - \\

& \small MultiCM \cite{shiba2022secrets} 
& 0.30 & 0.10 & \underline{0.42} & \underline{0.10} & \underline{0.60} & \underline{0.59} & \textbf{0.50} & \underline{0.28}
& \underline{0.46} & \underline{0.27} \\

\midrule

USL & \small FireFlowNet \cite{paredes2021back} 
& 1.06 & 6.60 & 0.97 & 2.60 & 1.67 & 15.30 & 1.43 & 11.00
& 1.28 & 8.88 \\

& \small EV-MGRFlow \cite{zhuang2024ev} 
& 0.28 & \underline{0.02} & 0.41 & 0.17 & 0.70 & 2.35 & 0.59 & 1.29
& 0.50 & 0.96 \\

& \small Zhu et al. \cite{zhu2019unsupervised} 
& 0.32 & \textbf{0.00} & 0.58 & \textbf{0.00} & 1.02 & 4.00 & 0.87 & 3.00
& 0.70 & 1.75 \\

\midrule

SSL & \small EV-FlowNet \cite{zhu2018ev} 
& 0.49 & 0.20 & 1.03 & 2.20 & 1.72 & 15.10 & 1.53 & 11.90
& 1.19 & 7.35 \\

& \small STE-FlowNet \cite{ding2022spatio} 
& 0.42 & \textbf{0.00} & 0.57 & \underline{0.10} & 0.79 & 1.60 & 0.72 & 1.30
& 0.63 & 0.75 \\

\midrule

SL & \small Matrix-LSTM \cite{cannici2020differentiable} 
& - & - & 0.82 & 0.53 & 1.19 & 5.59 & 1.08 & 4.81
& 1.03 & 3.64 \\

& \small E-RAFT \cite{eraft} 
& 0.24 & \textbf{0.00} & 1.10 & 5.72 & 1.94 & 30.79 & 1.66 & 25.20
& 1.24 & 15.43 \\

& \small DCEIFlow \cite{wan2022learning} 
& \underline{0.22} & \textbf{0.00} & 0.75 & 1.55 & 0.90 & 2.10 & 0.80 & 1.77
& 0.67 & 1.36 \\

& \small EDCFlow \cite{liu2025edcflow} 
& 0.23 & \textbf{0.00} & 0.97 & 2.19 & 1.49 & 10.79 & 1.38 & 8.98
& 1.02 & 5.49 \\

& \small TMA \cite{liu2023tma} 
& 0.25 & 0.07 & 1.06 & 3.63 & 1.81 & 27.29 & 1.58 & 23.26
& 1.18 & 13.56 \\

\midrule

& \small Ours
& \textbf{0.21} & 0.04 & \textbf{0.40} & \textbf{0.00} & \textbf{0.55} & \textbf{0.30} & \textbf{0.50} & \textbf{0.14}
& \textbf{0.42} & \textbf{0.12} \\

\bottomrule
\end{tabular}
}
\end{table*}

\begin{figure*}[t]
    \centering
    \includegraphics[width=0.7\textwidth]{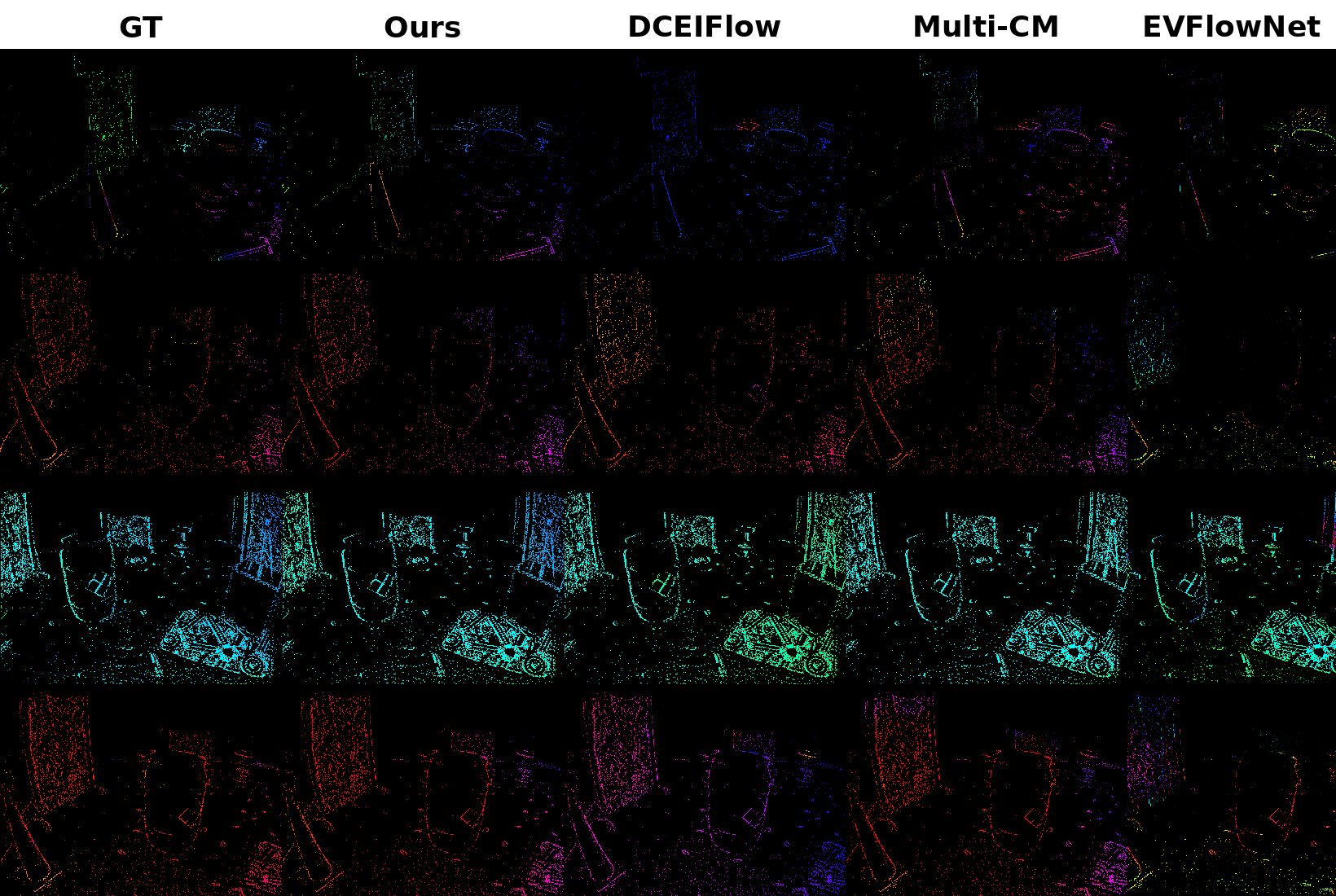}
    \vspace{-4pt}
    \caption{Qualitative results on the MVSEC Indoor Flying 3 scene. From left to right: ground truth, ours, DCEIFlow \cite{wan2022learning}, Multi-CM \cite{shiba2022secrets}, and EVFlowNet \cite{zhu2018ev}. Visualization is shown only at event-active pixels (i.e., pixels that received events within the evaluation interval).}
    \label{fig:qualitative}
\end{figure*}

\begin{figure*}[t]
  \centering
  \includegraphics[width=0.8\textwidth]{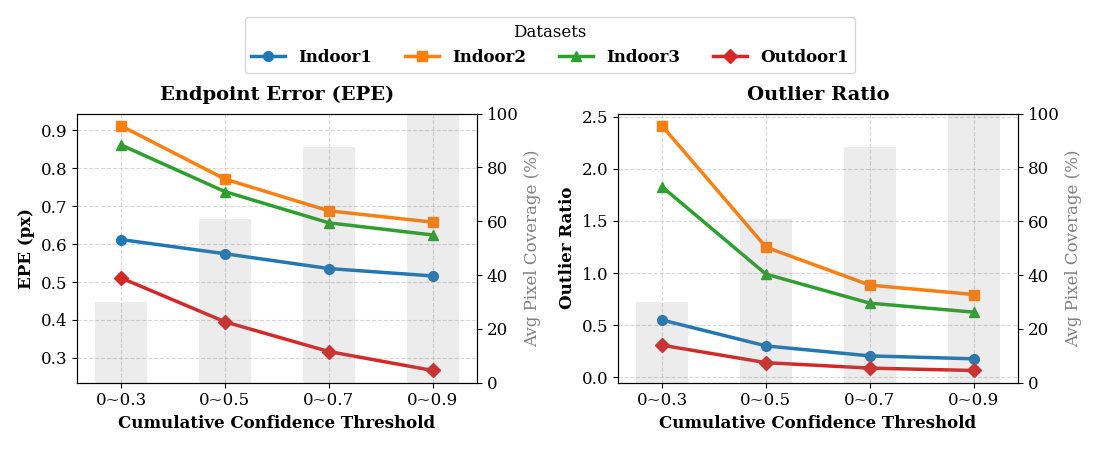}
  \caption{EPE and Outlier Ratio as a function of the confidence threshold on MVSEC sequences. Gray bars indicate pixel coverage. Errors decrease monotonically as the threshold increases, indicating that the learned confidence correlates with prediction quality.}
  \label{fig:conf_effect}
\end{figure*}

\subsection{Results}

\noindent\textbf{Local flow estimation.}
We first evaluate the Continuous Local Recurrent Network against recent local and incremental methods (\cref{tab:unified_local_flow_final_v6}). To measure intrinsic accuracy, we deploy the local estimator densely at all pixel locations, independent of the striding and aggregation used in the aggregation pipeline. LC-Flow outperforms existing baselines across all MVSEC sequences, achieving an EPE of $0.24$ on Outdoor Day 1 and $0.58$ on Indoor Flying 3.
Although trained to predict full 2D flow, LC-Flow attains lower PEE than VecKMFlow, which targets normal flow, suggesting that accurate full-flow predictions also capture normal-flow components. For a fair comparison with TEGBP at its reported evaluation interval, we additionally evaluate at $\Delta t=0.05$, where LC-Flow also outperforms TEGBP across all sequences.

\noindent\textbf{Global flow estimation via confidence-guided aggregation.}
Next, we compare LC-Flow with confidence-guided aggregation against existing optical flow methods (\cref{MVSECflow_table_withavg}). While supervised dense methods perform well on the training domain, their performance degrades substantially on unseen indoor sequences. In contrast, LC-Flow achieves the best EPE across all MVSEC sequences, with the lowest average EPE ($0.42$) and outlier ratio ($0.14\%$), demonstrating strong cross-domain generalization despite being trained on a single outdoor sequence. Qualitative comparisons on MVSEC Indoor Flying 3 (\cref{fig:qualitative}) further show that LC-Flow produces accurate and spatially coherent flow estimates closely aligned with the ground truth, whereas competing methods exhibit noisier or less accurate predictions.

\noindent\textbf{Zero-shot transfer to DSEC.}
We further apply the MVSEC-trained models to three DSEC sequences without fine-tuning (\cref{tab:dsec}). Since DSEC provides ground truth over a much longer interval ($100$\,ms) than the $\sim$22\,ms interval used in training, the predicted displacements are at a mismatched scale, and we do not evaluate absolute magnitude. Instead, we report a scale-aligned EPE (EPE$^\ast$), where a single scale factor per sequence and method is fitted to minimize EPE, together with the magnitude-independent angular error (AE). Under this protocol, LC-Flow with confidence-guided aggregation attains the lowest EPE$^\ast$ and AE on all three sequences (average $1.56$ / $8.6^\circ$). The local estimator alone reaches an EPE$^\ast$ comparable to the baselines but a higher AE, as purely local direction estimates remain aperture-limited; aggregation propagates reliable directions and reduces AE from $21.4^\circ$ to $8.6^\circ$. This suggests that the flow direction and spatial structure learned from local events transfer across datasets.%

\begin{table}[t]
\centering
\small
\caption{Zero-shot transfer from MVSEC to DSEC. Learning-based methods are trained on MVSEC and evaluated without fine-tuning; Multi-CM is optimization-based. EPE$^\ast$: scale-aligned EPE, with a single scale factor fitted per sequence and method; AE: angular error in degrees. $^\S$Images used in training; $^\dagger$images used in training \emph{and} inference.}
\label{tab:dsec}
\begin{tabular}{@{}l@{\hspace{5pt}}
                c@{\hspace{3pt}}c@{\hspace{8pt}}
                c@{\hspace{3pt}}c@{\hspace{8pt}}
                c@{\hspace{3pt}}c@{}}
\toprule
& \multicolumn{2}{c}{thun} & \multicolumn{2}{c}{zc11} & \multicolumn{2}{c}{zc05} \\
\cmidrule(lr){2-3}\cmidrule(lr){4-5}\cmidrule(lr){6-7}
Method & EPE$^\ast$ & AE & EPE$^\ast$ & AE & EPE$^\ast$ & AE \\
\midrule
EV-FlowNet$^\S$~\cite{zhu2018ev}          & 4.51 & 47.2 & 4.17 & 50.0 & 6.24 & 50.0 \\
E-RAFT~\cite{eraft}                       & 1.74 & 11.5 & 2.07 & 22.8 & 3.43 & 20.0 \\
DCEIFlow$^\dagger$~\cite{wan2022learning} & 2.35 & 12.3 & 1.43 & 11.3 & 2.30 & 12.8 \\
Multi-CM~\cite{shiba2022secrets}          & 1.84 & 15.1 & 2.17 & 23.0 & 3.56 & 27.5 \\
\midrule
Ours (Local)                              & 2.29 & 22.2 & 2.00 & 21.4 & 2.25 & 20.7 \\
Ours (Global) & \textbf{1.49} & \textbf{7.5} & \textbf{1.30} & \textbf{8.0} & \textbf{1.90} & \textbf{10.2} \\
\bottomrule
\end{tabular}
\end{table}

\begin{table}[tbp]
  \centering
  \small
  \caption{Ablation on training strategy and inference reset interval on MVSEC \textit{indoor\_flying3}. $M$ denotes the number of consecutive training slices, fixed at $1$ or randomly sampled from $[1,10]$ (Var.). ``Every $k$'' resets the GRU hidden state every $k$ inference slices. }
  \label{tab:continuous_reset}
  \resizebox{\columnwidth}{!}{%
  \begin{tabular}{l|ccccc}
    \toprule
    \multirow{2}{*}{\textbf{Training}}
      & \multicolumn{5}{c}{\textbf{Inference reset interval}} \\
    \cmidrule(lr){2-6}
      & Every 1 & Every 2 & Every 4 & Every 8 & No reset \\
    \midrule
    $M=1$
      & 1.17 & 1.33 & 1.98 & 2.82 & 4.19 \\
    Var. $M\in[1,10]$
      & 1.24 & 1.06 & 0.88 & 0.73 & \textbf{0.58} \\
    \bottomrule
  \end{tabular}
  }
\end{table}

\begin{table}[t]
\centering
\small
\caption{Ablation of the confidence-guided aggregator on MVSEC. Each configuration incrementally adds a component to the preceding one.}
\label{tab:ablation_fusion_comprehensive}
\begin{tabular}{c l cc}
\toprule
Exp. & Configuration & EPE ($\downarrow$) & \%Out ($\downarrow$) \\
\midrule
\#0 & Bilinear interpolation & 0.56 & 0.38 \\
\#1 & + Spatial aggregation & 0.54 & 0.26 \\
\#2 & + Multi-scale aggregation & 0.53 & 0.28 \\
\#3 & + Confidence-guided averaging & 0.51 & 0.18 \\
\#4 & + Learnable fusion & \textbf{0.50} & \textbf{0.14} \\
\bottomrule
\end{tabular}
\end{table}

\subsection{Effect of Confidence}
\begin{figure}[t]
    \centering
    \includegraphics[width=0.85\columnwidth]{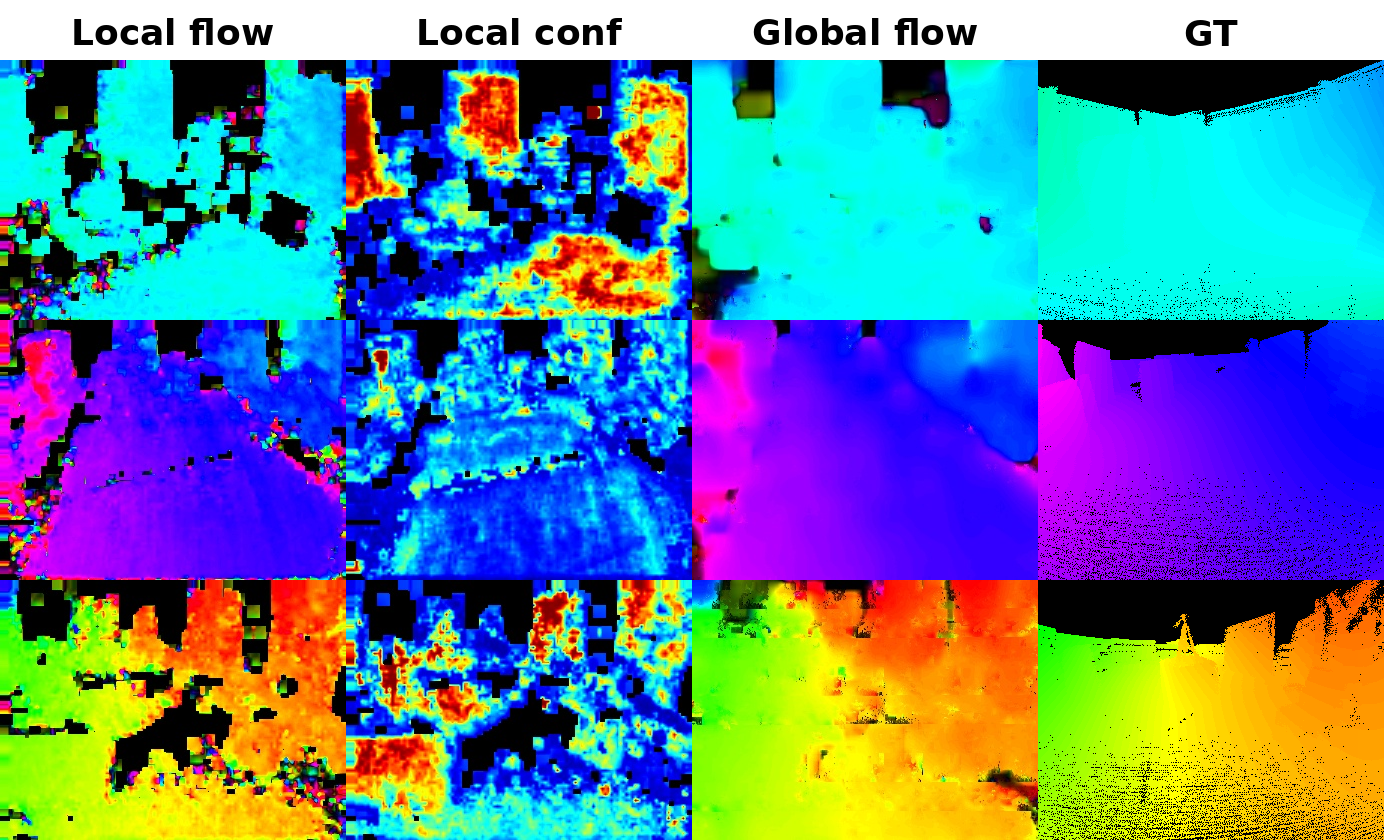}
    \caption{From left to right: Local flow, local confidence, confidence-guided aggregated flow, and ground truth on MVSEC indoor sequences.}
    \label{fig:pipeline_3scenes}
\end{figure}

We analyze the correlation between predicted confidence and actual error across MVSEC sequences. As shown in \cref{fig:conf_effect}, all error metrics decrease monotonically as the confidence threshold increases, across both indoor and outdoor sequences, confirming that the learned confidence reliably reflects prediction quality. \cref{fig:pipeline_3scenes} shows how the sparse local flow and confidence are integrated into a globally consistent field, improving spatial coverage and coherence over the local outputs alone.

\subsection{Ablation Studies}

\subsubsection{Temporal Persistence and Continuous Training}
We compare a model trained on single slices ($M=1$) with our continuous training strategy using randomly concatenated sequences ($M\sim\mathcal{U}[1,10]$), under different inference reset intervals. During inference, we reset the GRU hidden state every $R$ temporal slices and also report the continuous \emph{no-reset} setting.
As shown in \cref{tab:continuous_reset}, the single-slice model ($M=1$) degrades sharply as $R$ increases (EPE $1.17\rightarrow4.19$), indicating instability when the recurrent state is carried over long durations without appropriate training. In contrast, the multi-slice model improves with longer context and achieves its best performance in the no-reset setting (EPE $0.58$). While $M=1$ is slightly better at the shortest reset interval, it fails to generalize to continuous streaming, highlighting the need for persistent states trained with multi-slice sequences.

\subsubsection{Confidence-guided Aggregator}
We ablate the aggregation pipeline used to generate global flow, as shown in \cref{tab:ablation_fusion_comprehensive}. Starting from bilinear interpolation of sparse predictions (\#0), spatial neighborhood averaging (\#1) improves robustness, and multi-scale fusion with uniform averaging (\#2) further helps by combining fine and coarse contexts. Confidence-weighted aggregation (\#3) yields additional gains by suppressing unreliable estimates, and a lightweight CNN for per-scale weights (\#4) provides the best reconstruction by adaptively selecting informative scales.

\subsubsection{Operating Frequency and Latency}

We measure latency across operating frequencies on MVSEC with RTX A6000 as described in \cref{tab:freq_latency}. Since the chunk size only schedules computation, the recurrent state evolution, and hence the predicted flow, is identical across $45$, $100$, and $200$\,Hz, while latency decreases with fewer events per interval. The confidence-averaging configuration (\#3) stays within the cycle budget at every tested frequency, whereas the full aggregator (\#4) trades latency for accuracy. A runtime comparison with baselines is given in the supplementary material.

\begin{table}[tbp]
  \centering
  \small
  \setlength{\tabcolsep}{6pt}
  \caption{Latency versus operating frequency on MVSEC \textit{outdoor\_day1}, measured on RTX A6000 ($s{=}3$). Because latency scales with the number of events per query interval, the real-time configuration (\#3) stays below the cycle budget at every frequency, while the most accurate variant (\#4) remains above it.}
  \label{tab:freq_latency}
  \begin{tabular}{c c c c c}
    \toprule
    Freq. & Cycle & \multicolumn{3}{c}{Latency (ms)} \\
    \cmidrule(lr){3-5}
    (Hz) & (ms) & GRU only & \#3 & \#4 \\
    \midrule
    45  & 22.2 & 16.83 & \textbf{16.66} & 41.93 \\
    100 & 10.0 &  8.26 &  \textbf{8.54} & 33.48 \\
    200 &  5.0 &  4.70 &  \textbf{4.96} & 29.66 \\
    \bottomrule
  \end{tabular}
\end{table}

\section{Limitations and Future Work}
LC-Flow requires flow supervision over short intervals, a mild condition that is readily satisfied in high-frequency event-camera settings; it is constrained when ground truth is only available exclusively over long intervals. The learned confidence is also used as a relative reliability score rather than a calibrated uncertainty, and a calibration analysis relating predicted confidence to empirical error is a natural extension. Two failure regimes bound its scope: locally ambiguous regions (weak or aperture-limited texture), where insufficient local evidence yields low confidence rather than silent errors; and large or non-linear motion, where the regressed endpoint error grows with displacement. Looking ahead, the sparse local flow and confidence produced by LC-Flow are well suited to downstream motion tasks. In particular, we plan to integrate them into an event-based visual odometry pipeline, where confidence-filtered motion cues can directly support tracking and pose estimation. More broadly, validating LC-Flow across additional sensors and scene types would further establish its generality.

\section{Conclusion}
We presented LC-Flow, a learning-based framework for temporally continuous event-based optical flow estimation from local event dynamics. LC-Flow maintains persistent hidden states per local grid to support event-driven, continuous-time flow querying at arbitrary timestamps. By jointly predicting flow and confidence, it mitigates ambiguity from sparse local observations and supports confidence-guided reconstruction of a full-resolution flow field. Because it relies only on local motion, LC-Flow generalizes across domains: trained on a single outdoor sequence, it remains state-of-the-art on indoor scenes, and its confidence-guided aggregation attains overall state-of-the-art on MVSEC. These results suggest that accurate local dynamics with explicit reliability estimation offer a domain-robust alternative to heavy global, dense architectures.

\section{Acknowledgment}
This work was supported by National IT Industry Promotion Agency(NIPA) grant funded by the Korea government(MSIT) (RS-2026-25621677).

{
    \small
    \bibliographystyle{ieeenat_fullname}
    \bibliography{main}
}

\clearpage
\makeatletter
\twocolumn[{%
   \null
   \vskip .375in
   \begin{center}
      {\Large \bf Supplementary Material for\\ LC-Flow: Learning Local Continuous Optical Flow and Confidence from Events \par}
      \vspace*{24pt}{
        \large
        \lineskip .5em
        \begin{tabular}[t]{c}
Gunwoo Jeon$^{1}$ \quad Chaesong Park$^{1}$ \quad Jongwoo Lim$^{1,2}$\\
IPAI$^{1}$ \& ME$^{2}$, Seoul National University\\
{\tt\small \{gunw0, chase121, jongwoo.lim\}@snu.ac.kr}
        \end{tabular}
        \par
      }
      \vskip .5em
      \vspace*{12pt}
   \end{center}
}]
\makeatother
\setcounter{section}{0}
\setcounter{figure}{0}
\setcounter{table}{0}
\setcounter{equation}{0}
\renewcommand{\theHsection}{supp.\arabic{section}}
\renewcommand{\theHsubsection}{supp.\arabic{section}.\arabic{subsection}}
\renewcommand{\theHfigure}{supp.\arabic{figure}}
\renewcommand{\theHtable}{supp.\arabic{table}}
\renewcommand{\theHequation}{supp.\arabic{equation}}

\section{Qualitative Result Video on MVSEC}
\label{supp:sec:qual}

We provide a supplementary video demonstrating qualitative results on
the MVSEC \textit{outdoor\_day1} sequence. Each frame shows, from left to right, local flow,
confidence map, confidence-guided aggregated global flow, and ground-truth
flow. In the confidence map, red indicates high confidence and blue
indicates low confidence. Regions with sparse event activity produce
blocky local flow estimates, which are correctly assigned low confidence
(blue); these regions are then suppressed during aggregation, resulting
in a globally consistent flow field. The video illustrates that this
behavior is temporally stable across the full sequence.

\section{Confidence Balancing Coefficient \boldmath$\lambda$}
\label{supp:sec:lambda}

The confidence balancing coefficient $\lambda$ controls the degree of
confidence regularization in the loss (Eq.~3 of the main paper).
A large $\lambda$ encourages the network to assign uniformly high
confidence across all regions, resulting in near-uniform confidence
maps that lose discriminative information (visible for $\lambda = 0.4$
and $\lambda = 0.8$, where the maps become saturated with high values).
Conversely, a small $\lambda$ drives confidence towards lower values.
We sweep $\lambda \in \{0.1, 0.2, 0.4, 0.8\}$ on MVSEC and select
$\lambda = 0.2$ as the operating point, which produces confidence maps
that are spatially discriminative.

\begin{figure}[t]
    \centering
    \includegraphics[width=\columnwidth]{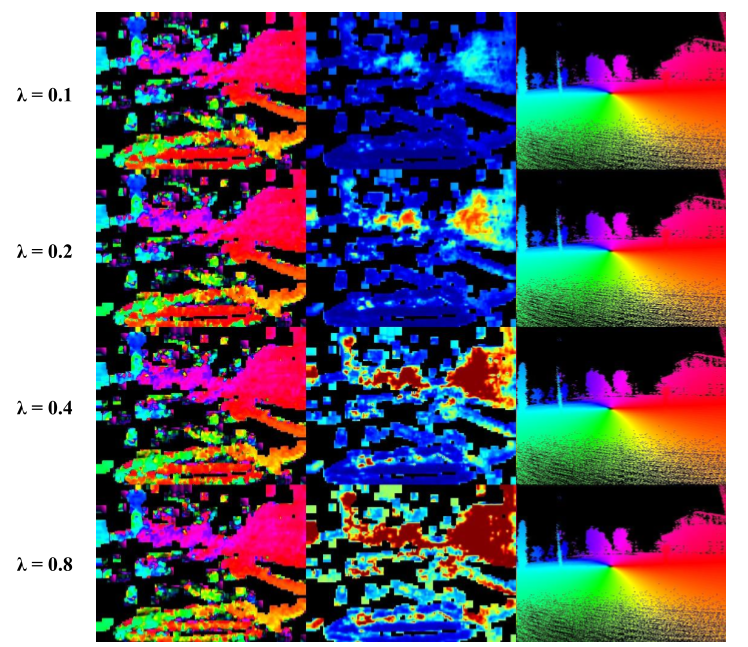}
    \caption{Qualitative visualization of flow and confidence under
    different $\lambda$ values on MVSEC.
    Rows correspond to $\lambda = 0.1, 0.2, 0.4, 0.8$ from top to bottom;
    columns show local flow, confidence map, and ground-truth
    flow from left to right.
    At $\lambda = 0.4$ and $\lambda = 0.8$, the confidence maps become
    near-uniform (saturated), losing the ability to discriminate reliable
    from unreliable predictions.}
    \label{supp:fig:lambda_ablation}
\end{figure}

\section{Network Architecture Details}
\label{supp:sec:arch}

We summarize the architecture of the prediction heads and the
Confidence-guided Aggregator in \cref{supp:tab:arch}.
The confidence head uses a deeper MLP than the flow head to capture
the more complex mapping from hidden state to reliability score.
In terms of memory, the per-grid state is a single hidden vector ($H{=}256$, fp32; $1$\,KB), so the persistent-state footprint is $N_{\text{active}}\times H\times4$\,B: $\approx88$\,MB for MVSEC ($346\times260$) and $\approx300$\,MB for DSEC ($640\times480$) at full resolution. Empty grids hold no state, and stride $s$ reduces $N_{\text{active}}$ by $\approx s^2$ ($\approx10$\,MB on MVSEC at $s{=}3$), giving a direct memory knob for higher-resolution deployment.

\begin{table}[t]
\centering
\caption{Architecture of the flow and confidence prediction heads
in the Local Estimator, and configuration of the Confidence-guided
Aggregator.}
\label{supp:tab:arch}
\setlength{\tabcolsep}{4pt}
\small
\begin{tabular}{p{1.9cm}p{1.7cm}p{3.6cm}}
\toprule
\textbf{Module} & \textbf{Component} & \textbf{Architecture} \\
\midrule
\multirow{2}{1.9cm}{Local Estimator}
& Flow head & FC(256{\small$\to$}64) $\to$ ReLU $\to$ FC(64{\small$\to$}2) \\[2pt]
& Conf head & 6-layer FC (256{\small$\to$}1) $\to$ Sigmoid \\
\midrule
\multirow{5}{1.9cm}{Conf-guided Aggregator}
& Scales $s$     & $\{1,2,4\}$ \\[2pt]
& Patch size $k$ & 7 \\[2pt]
& CNN       & Conv(9{\small$\to$}32{\small$\to$}64) $\to$ AdaptAvgPool \\
&                & $\to$ FC(64{\small$\to$}32{\small$\to$}3) $\to$ Softmax \\[2pt]
& LR / Epochs    & $10^{-4}$ / 50 \\[2pt]
& Batch size     & 4 \\
\bottomrule
\end{tabular}
\end{table}

\section{Runtime Comparison}
\label{supp:sec:speed}

We report the speed and accuracy of LC-Flow configurations and baselines in \cref{supp:tab:speed_accuracy}; all timings are measured on a single RTX A6000. Variants \#0--\#4 follow the aggregator ablation of the main paper (\#0: bilinear interpolation; \#3: confidence-weighted multi-scale averaging; \#4: full confidence-guided aggregation with the patch-based CNN). At a fixed $45$\,Hz, the real-time configuration (\#3, $s{=}5$) runs at $8.53$\,ms with an EPE of $0.21$, about $2\times$ faster than E-RAFT ($17.24$\,ms) and also more accurate ($0.24$ EPE); the most accurate variant (\#4) is slower ($40.67$\,ms) due to the patch-based CNN. 

\begin{table}[tbp]
  \centering
  \small
  \setlength{\tabcolsep}{5pt}
  \caption{Effect of stride and aggregation variant on speed and accuracy,
  evaluated on MVSEC \textit{outdoor\_day1} (45Hz).}
  \label{supp:tab:speed_accuracy}
  \scalebox{0.90}{%
  \begin{tabular}{l c c}
    \toprule
    Method & Inference Time & EPE \\
           & (ms/frame) & ($\downarrow$) \\
    \midrule
    Multi-CM~\cite{s:shiba2022secrets}   & $>$8000 & 0.30 \\
    E-RAFT~\cite{s:eraft}                & 17.24   & 0.24 \\
    VecKM~\cite{s:Veckmflow} & 83.24   & $0.88^\dagger$ \\
    \midrule
    \multicolumn{3}{l}{LC-Flow (Ours)} \\
    \quad \#0 Bilinear (s=3)            & 16.49 & 0.23 \\
    \quad \#4 Conf-guided Agg (s=3)     & 40.67 & 0.21 \\
    \quad \#3 Conf Avg (s=3)            & 16.66 & 0.21 \\
    \quad \#3 Conf Avg (s=4)            & 11.18 & 0.21 \\
    \quad \#3 Conf Avg (s=5)            &  8.53 & 0.21 \\
    \bottomrule
    \multicolumn{3}{l}{\footnotesize $^\dagger$ PEE; a lower bound of EPE.}
  \end{tabular}
  }
\end{table}

\section{Visual Odometry Application}
\label{supp:sec:vo}

We validate the practical utility of LC-Flow by integrating its
sparse local flow and confidence into SMF-VO~\cite{s:yang2025smf},
a sparse optical flow-based visual odometry pipeline.

\subsection{Setup}

We evaluate on the MVSEC \textit{indoor\_flying3} sequence at
$\Delta t = 0.05$\,s and compare three flow inputs:
(i)~ground-truth flow (upper bound),
(ii)~DCEIFlow~\cite{s:wan2022learning} (supervised baseline), and
(iii)~our sparse local flow, with and without confidence-based filtering
(thresholds 0.0 and 0.3, respectively).
The local estimator is trained exclusively on the \textit{outdoor\_day2}
sequence and applied directly to the indoor scene without fine-tuning,
following the standard cross-domain evaluation protocol.

\subsection{Results}

\begin{figure*}[tb]
    \centering
    \begin{subfigure}[b]{0.42\linewidth}
        \centering
        \includegraphics[width=\linewidth]{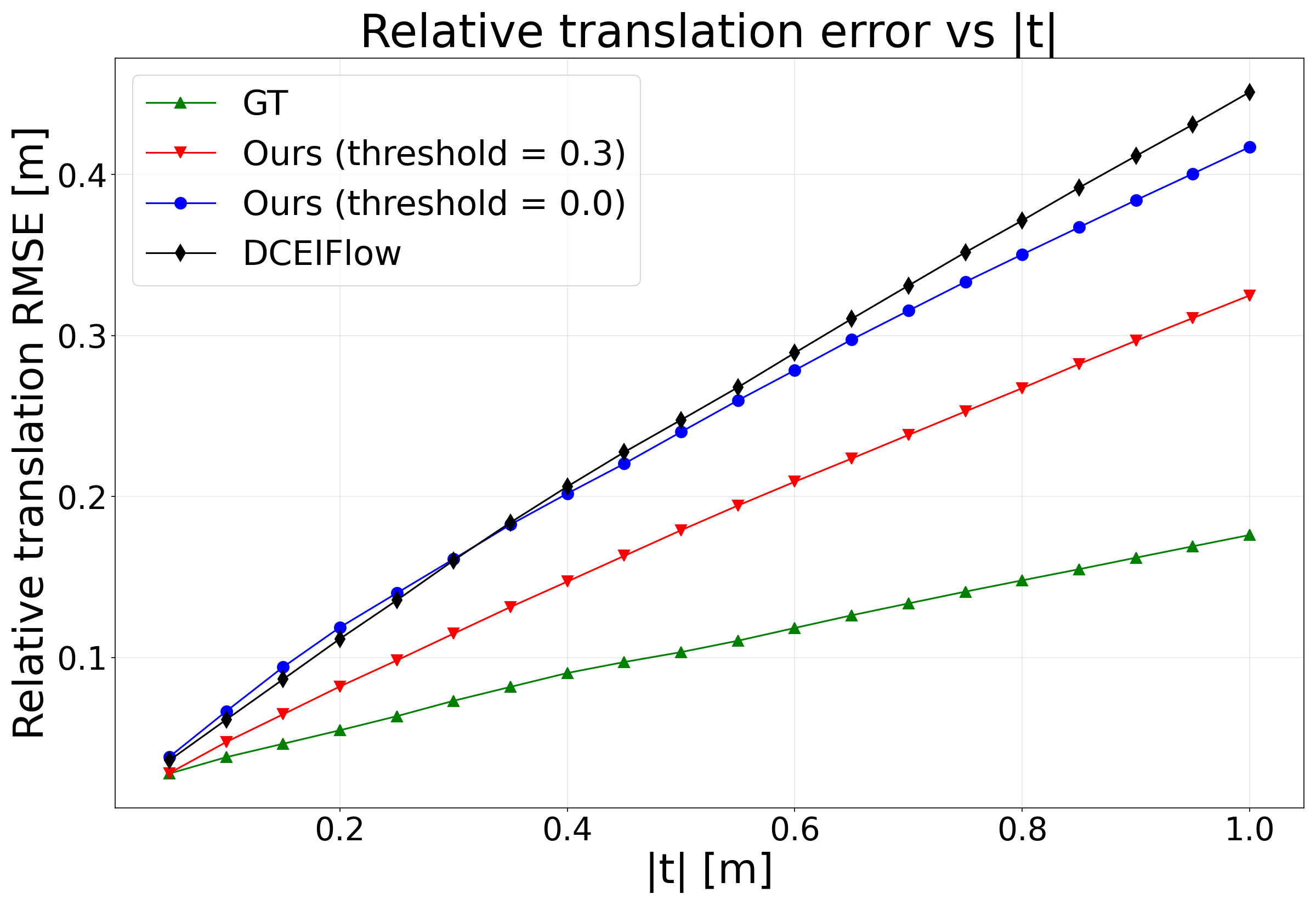}
        \caption{Translation Error}
        \label{supp:fig:translation_error}
    \end{subfigure}
    \hfill
    \begin{subfigure}[b]{0.42\linewidth}
        \centering
        \includegraphics[width=\linewidth]{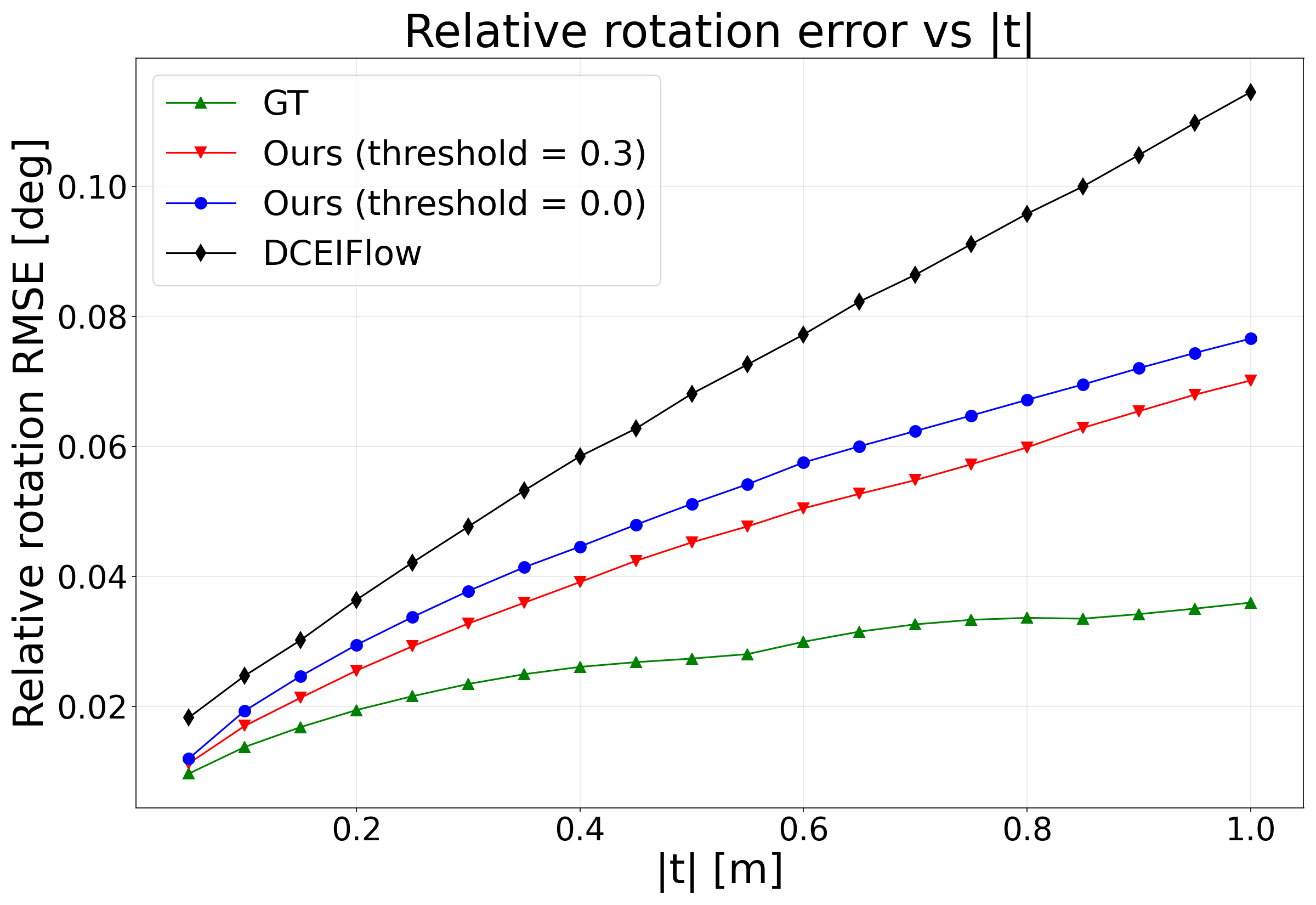}
        \caption{Rotation Error}
        \label{supp:fig:rotation_error}
    \end{subfigure}
    \caption{Relative RMSE as a function of translation magnitude $|t|$ on MVSEC \textit{indoor\_flying3}.
    Both plots show that our method outperforms DCEIFlow.
    (a) Applying a stricter confidence threshold (0.3 vs.\ 0.0) consistently reduces translation error, confirming that learned confidence identifies high-quality flow.
    (b) The benefit of confidence filtering is less pronounced for rotation, as rotational motion is inherently more challenging.}
    \label{supp:fig:combined_errors}
\end{figure*}

\cref{supp:fig:translation_error,supp:fig:rotation_error} report relative
translation and rotation RMSE as a function of translation magnitude
$|t|$.
Applying a stricter confidence threshold (0.3 vs.\ 0.0) consistently
reduces translation error across all motion magnitudes, with the gap
growing as $|t|$ increases, confirming that the learned confidence
reliably identifies high-quality flow estimates.
The effect on rotation is more modest, as rotational motion is
inherently harder to resolve from local observations alone.
In both metrics, our method outperforms DCEIFlow, which relies on
image frames in addition to events for dense flow estimation.

\end{document}